\documentclass[preprint,12pt]{elsarticle}
\usepackage[pdfborder={0 0 0}]{hyperref}
\usepackage{amsthm,amsmath,amsfonts,latexsym,amssymb}
\usepackage{graphics,fancyhdr,graphicx}
\usepackage[ruled,linesnumbered]{algorithm2e}
\usepackage[margin=1in]{geometry}
\usepackage{multirow,booktabs}
\usepackage[table]{xcolor}
\usepackage{listings}
\usepackage{tabularx}
\usepackage{placeins}
\usepackage{comment}
\usepackage{lscape}
\usepackage{bm}
\usepackage{hyperref}
\usepackage{mathrsfs}
\usepackage{arydshln}
\usepackage{appendix}
\usepackage{soul}
\definecolor{custom-blue}{RGB}{3,69,173}
\hypersetup{
    colorlinks = true,
    urlcolor   = blue,
    citecolor  = custom-blue,
}
\biboptions{numbers,sort&compress}
\definecolor{listinggray}{gray}{0.9}
\definecolor{lbcolor}{rgb}{0.9,0.9,0.9}
\definecolor{Darkgreen}{RGB}{0,100,0}
\usepackage[font=small,labelfont=bf, textfont=it]{caption}

\usepackage{float}
\usepackage{booktabs}
\usepackage{array}
\usepackage{multirow}
\usepackage{tabularray}
\usepackage{hhline}
\usepackage{algpseudocode}
\usepackage{mathtools}
\usepackage{subcaption}
\usepackage{enumitem}
\usepackage{lipsum}

\makeatletter
\def\ps@pprintTitle{%
 \let\@oddhead\@empty
 \let\@evenhead\@empty
 \def\@oddfoot{}%
 \let\@evenfoot\@oddfoot}
\makeatother

\begin{document}
\abovedisplayskip=6.0pt
\belowdisplayskip=6.0pt


\newcommand{\Review}[2][blue]{{\textcolor{#1}{#2}}}

\begin{frontmatter}

\title{Learning Nonlinear Responses in PET Bottle Buckling with a Hybrid DeepONet–Transolver Framework}

\author[1]{Varun Kumar}
\ead{varun_kumar2@brown.edu}
\author[2]{Jing Bi}
\author[2]{Cyril Ngo Ngoc}
\author[2]{Victor Oancea}
\author[1,3]{George Em Karniadakis\corref{cor1}}
\ead{george_karniadakis@brown.edu}

\address[1]{School of Engineering, Brown University}
\address[2]{
Dassault Syst\`emes}
\address[3]{Division of Applied Mathematics, Brown University}

\cortext[cor1]{Corresponding author.}

\begin{abstract}
\noindent
Neural surrogates and operator networks for solving partial differential equation (PDE) problems have attracted significant research interest in recent years. However, most existing approaches are limited in their ability to generalize solutions across varying non-parametric geometric domains. In this work, we address this challenge in the context of Polyethylene Terephthalate (PET) bottle buckling analysis, a representative packaging design problem conventionally solved using computationally expensive finite element analysis (FEA). We introduce a hybrid DeepONet-Transolver framework that simultaneously predicts nodal displacement fields and the time evolution of reaction forces during top load compression. Our methodology is evaluated on two families of bottle geometries parameterized by two and four design variables. Training data is generated using nonlinear FEA simulations in Abaqus for 254 unique designs per family. The proposed framework achieves mean relative $L^{2}$ errors of 2.5–13\% for displacement fields and approximately 2.4\% for time-dependent reaction forces for the four-parameter bottle family. Point-wise error analyses further show absolute displacement errors on the order of $10^{-4}$–$10^{-3}$, with the largest discrepancies confined to localized geometric regions. Importantly, the model accurately captures key physical phenomena, such as buckling behavior, across diverse bottle geometries. These results highlight the potential of our framework as a scalable and computationally efficient surrogate, particularly for multi-task predictions in computational mechanics and applications requiring rapid design evaluation.

 \end{abstract}
\end{frontmatter}

\section{Introduction}
\label{sec:intro}
Developing neural surrogates to learn functional mappings for PDE  problems defined on non-Euclidean domains has recently gained significant research attention. Traditional machine learning tasks in fields such as computer vision and signal processing benefit from statistical properties like stationarity and compositionality,  which naturally arise from the Euclidean structure of the data \cite{Geometric_DL_Bronstein}. By contrast, engineering design problems frequently involve non-Euclidean data, where characteristics such as common coordinate system or shift invariance do not exist. Consequently, standard machine learning architectures such as Convolutional Neural Networks (CNNs) cannot be directly applied to such problems. An equally important challenge lies in the effective representation of non-Euclidean geometry data, as it plays a central role in determining the quality of the input–output mapping. Common strategies for representating such geometries include voxels, point clouds, meshes, zero level-set function such as Signed Distance Fields (SDFs), and parametric encoding, each offering distinct advantages and limitations \cite{Deep_generative_learning_Faez}. 

Operator networks, such as Deep Operator Network (DeepONet) \cite{deeponet_lulu_2021}, Fourier Neural Operator (FNO) \cite{FNO_2020}, Wavelet Neural Operator (WNO) \cite{WNO}, among others, have emerged as powerful tools for neural PDE solving tasks. These approaches offer key advantages over conventional machine learning frameworks, including discretization invariance and improved generalizability. While these methods have demonstrated promising results across a range of applications, their applicability to learn PDE solutions consistently across multiple geometric domains remains limited. As a result, growing research efforts are directed towards developing \emph{geometry-aware} neural surrogates that can effectively  capture the mapping of PDE solutions over diverse geometries. 

In this work, we focus on the problem of PET bottle buckling, an important task in computational mechanics and packaging design. Traditionally, this problem is addressed using finite element analysis (FEA), where each candidate geometry must be meshed and solved under prescribed loading conditions. While accurate, this approach is computationally intensive, particularly when large design spaces need to be explored, such as in optimization or uncertainty quantification studies. Evaluating hundreds or thousands of bottle designs, each requiring a nonlinear FEA simulation with potentially thousands (or even millions) of degrees of freedom, is prohibitively expensive and restricts the pace of design iteration. To overcome these limitations, we propose a \emph{geometry-aware} neural surrogate that learns the mapping from bottle geometry to key mechanical responses, including displacement fields and time-dependent reaction forces. Specifically, we extend the Transolver framework \cite{transolver} by integrating it with DeepONet, enabling simultaneous prediction of spatial and temporal responses. While Transolver effectively maps geometric features to nodal displacements, it is inherently restricted to node-based feature learning and lacks the ability to predict time-dependent outputs. By integrating Transolver with DeepONet, we address this limitation and design a hybrid DeepONet-Transolver framework for handling multi-task learning in bottle buckling analysis.

The key contributions of this work are:

\begin{itemize}
    \item We develop a geometry-aware neural surrogate for bottle buckling analysis in computational mechanics and packaging design. The surrogate enables efficient evaluation of multiple bottle designs, thereby accelerating design exploration.
    \item We introduce a hybrid Transolver–DeepONet framework that extends the Transolver’s capability from learning spatial mappings alone to simultaneous prediction of nodal displacements and time-dependent reaction forces, making it applicable to broader classes of multi-task problems in computational mechanics.
\end{itemize}

\section{Background and related works}
\label{sec:Lit_review}
In this section, we review recent advances in \emph{geometry-aware} neural methods, organizing them according to the underlying geometry representation. In particular, we highlight approaches based on meshes, parametric representations, and point clouds, as these formats have emerged as versatile and widely adopted choices for non-Euclidean geometries. Their ability to capture complex geometrical features while balancing efficiency and accuracy makes them especially well-suited for the development of neural surrogates.

\vspace{5pt}
\noindent
\textbf{Methods using mesh-based representation} \\
Mesh-based representations are among the most widely used in engineering and scientific applications, and consequently, several neural surrogates have been developed to leverage this format. Geo-FNO \cite{Geo-FNO} extends the FNO framework to irregular meshes by introducing a mapping that converts irregular grids to regular grids before applying Fourier feature learning. While this ensures computational efficiency, it requires additional overhead for grid mapping, which becomes particularly challenging for complex engineering geometries. MeshGraphNets \cite{meshgraphnet} use Graph Convolutions to map the node and edge features to the PDE solution, exploiting the inherent connectivity information of meshes. However, graph-based methods are limited by their efficacy and effectiveness in learning multi-scale and long distance geometric relationships to physical quantities. X-MeshGraphNet \cite{Xmeshgraphnet_NVIDIA} improves the scalability of graph-based approach by partitioning large graphs into smaller sub-graphs and subsequently applying an aggregation technique to recover global solutions. This method has been successfully demonstrated for predicting aerodynamic pressure distributions and wall shear stress across 500 parametrically morphed car geometries in the DrivaerML dataset \cite{drivaerml}.

\vspace{5pt}
\noindent
\textbf{Methods using parametric representation} \\
Parametric representations provide a compressed, scalar encoding of geometric variations and have been used in geometry-based deep learning recently. For instance, \cite{airfoil_shukla} used a DeepONet framework to learn flow fields around different airfoil designs using two shape-defining parameters as inputs to the Branch network. In \cite{fusiondeeponet}, geometric parameters were used to model the hypersonic flow field around around two-dimensional objects to simulate capsule re-entry from outer space. Similarly, \cite{synergistic_learning} utilized geometric parameters for predicting the temperature distribution in multiple three-dimensional conductive plates. Geom-DeepONet \cite{geom_deeponet} combined parametric representation with applied load vector to predict stress fields in simple beam designs. Despite their success, these methods are limited to parametric geometries and cannot handle non-parametric designs where the key differentiating features are unknown. This limitation is significant in engineering practice, where geometries are often represented in generic formats (e.g., STEP or IGES) to ensure cross-platform compatibility and protect intellectual property.

\vspace{5pt}
\noindent
\textbf{Methods using point cloud representation} \\
Point clouds provide a memory-efficient representation of geometry. Although unstructured, the spatial arrangement of points can be exploited to encode geometric information in neural networks. Several prior works, such as \cite{pointnet, pointnet++, pointcnn,pointmlp} have been proposed to extract coordinate-based features from point clouds for computer graphic tasks such as classification and segmentation. More recently, these techniques have been adapted to develop neural surrogates for scientific regression tasks. For instance, \cite{mionet} proposed a DeepONet-based framework for predicting solution to Poisson's equation across varying PDE parameters, boundary conditions, forcing functions, and multiple convex polygon geometries. Drawing inspiration from PointNet, \cite{point_deeponet} introduced a DeepONet model that maps non-parametric bracket shapes, represented as point clouds, and variable loading conditions to stress fields resulting from linear elasticity. GINO \cite{GINO} extended the FNO framework to arbitrary grids by transforming point clouds to a uniform grid using a connectivity-based approach with a ball $B_r(x)$ around each grid point $x$, enabling discretization-invariant predictions. However, GINO incurs substantial pre-processing overhead, especially when applied to large domains.

CORAL \cite{CORAL} employs an encoder-decoder architecture, where the encoder produces a latent vector for input functions and measurement points via Implicit Neural Representation (INR) and the decoder maps this latent representation to outputs on different grids. CORAL demonstrated its effectiveness on a range of problems such as flow around two-dimensional airfoil designs, incompressible flow through pipes, and stress fields in a hyper-elastic material. Its key limitation is the requirement of an average deformation reference mesh with the same nodal points as the target mesh, which is often infeasible in practical engineering settings. DOMINO \cite{domino} addressed the need to capture finer, local geometric details using a sequence of point convolution layers to project geometry onto the whole computational domain. While this framework was successful in predicting aerodynamic flow around complex car geometries, the model exhibited relative errors of $\approx 20\%$ and requires significant pre-processing overhead.  Similarly, \cite{physics_geom_aware_Meidi} proposed a physics-informed DeepONet for multi-geometry, multi-parameter problems such as Darcy flow and linear elasticity. This approach encodes geometric boundaries using point-wise MLPs, which becomes computationally expensive for large-scale problems. Also, determining the boundaries in a complex three-dimensional shape is a challenging task, especially for geometries with fine feature variation between them.

Recently, Transformers \cite{attention_transformers} with their attention-based architectures, have opened a promising direction for developing \emph{geometry-aware} neural surrogates for scientific machine learning. For instance, GNOT \cite{GNOT} introduced a heterogeneous linear attention block to mitigate the quadratic computational cost of standard attention mechanism, enabling the integration of multiple inputs into a unified attention score. In addition, a gating mechanism was proposed to enhance scalability for large-scale problems. GNOT demonstrated strong performance across diverse applications, including transonic flow over different airfoil designs, linear elasticity in varying domains, and multi-cavity flow prediction. Transolver \cite{transolver} and its extension Transolver++ \cite{transolver++}, provide an alternative strategy for multi-geometry problems. Prior point cloud-based frameworks typically rely on MLPs and pooling layers to encode geometry, which often results in information loss - particularly when fine geometric variations are critical for design analysis. In contrast, Transolver uses physically-consistent tokens of spatial coordinates and employs point-wise spatial outputs as supervised signals to align token embeddings through local attention. This approach removes the need for a separate geometry encoder while improving predictive accuracy by leveraging attention across physically-consistent point features. Such attention-based methods have shown excellent scalability and robustness in large scale applications, such as DrivaerNet++ \cite{drivaernet++}. Other works based on Transformer attention such as Latent Neural Operator (LNO) \cite{latentneuraloperator} have also shown promising results in \emph{geometry-aware} problems. 

In this study, our objective is to leverage the local feature extraction capabilities of Transolver framework to learn the buckling characteristics of multiple packaging bottle designs under top compressive load. Specifically, for a set of non-parametric domain geometries $\Omega_i$, we aim to approximate the operator $\mathcal{G}$ 
\begin{equation}
    \mathcal{G}: \{\Omega_i\} \longrightarrow u_{i} (\mathbf{x}), \quad  \mathbf{x} \in \Omega_{i}, \quad \Omega_{i} \in \mathbf{\Omega} \subseteq \mathbb{R}^3, 
\end{equation}
 where $u_{i}(\mathbf{x})$ denotes the solution at the set of collocation points, $\{x_j\}_{j=1}^{n_i}$, that define the geometry $\Omega_{i}$. In addition to spatial field prediction, we also seek to predict the transient reaction force $F_{R}$, experienced during bottle compression, which is a measure of the bottle stiffness and its buckling characteristics. Since Transolver only predicts spatial fields at collocation points $x_{j}$, we integrate a DeepONet framework to concurrently predict reaction force $F_R(t)$. Specifically, DeepONet learns the operator
 \begin{equation}
     \mathcal{H}_{\Theta}: \mathcal{K}(\Omega_{i}) \times T \longrightarrow \mathbb{R}, \quad (k, t) \mapsto F_R(t), t\in [0,1], k\in\mathbb{R}^{1 \times D},
 \end{equation}
 where $\mathcal{K} \subseteq \mathbb{R}^{1 \times D}$ denotes the features extracted from the Transolver model during training. 

\section{Problem setting} 
\label{sec:prob_def}
Plastic bottle design is a critical problem in consumer packaging, where the competing objectives of material reduction and structural performance must be simultaneously satisfied. While minimizing material usage helps reduce environmental costs of using plastic, bottle designs must maintain sufficient strength to withstand stresses experienced in their manufacturing process and transportation. A key requirement for packaging bottles is their ability to resist excessive deformation during top loading which occurs in storage. Compressive forces exerted during such top loading conditions can induce structure collapse or buckling in bottle. Thus, careful geometry optimization is required to develop reliable bottle designs while minimizing plastic usage.

Traditional bottle design and evaluation follows an iterative process involving computer-aided design (CAD) and finite element analyses (FEA) where candidate designs are generated using CAD, then simulated with FEA and refined until the design objectives are met. This process is computationally expensive, especially when the design space is large since each design evaluation requires full FEA simulation before refinement. Neural surrogates offer a compelling alternative in such scenario since they can generate simulation outputs for multiple geometries almost instantaneously once the surrogate has been trained on historic data. Such an approach can help minimize product development time significantly, especially during concept exploration stage where large design spaces need to be explored rapidly with reasonable accuracy.

\subsection{Geometry selection}
\label{subsec:geom_selection}

To generate new bottle geometries for our analysis, we employed a parametric design approach that allows systematic variation of key geometric features. Two families of designs were considered, distinguished by the number of parameters varied: two-parameter and four-parameter designs. In the two-parameter family, the top section radius ($r_{\text{top}}$) and rib spacing ($d_{\text{rib}}$) in the upper section were varied to create distinct shapes within a Full Factorial design space $\mathcal{D}$
\begin{equation}
    \mathcal{D}:=[20, 35] \times [10, 25] \subset \mathbb{R}^2, \quad (r_{\text{top}}, d_{\text{rib}}) \in \mathcal{D}.
\end{equation}
The four-parameter design family adds more design variability by including two additional design parameters: rib radius ($r_{\text{rib}}$) and rib pitch ($p_{\text{rib}}$) such that
\begin{equation}
    \mathcal{D}:=[2, 4.5] \times [20, 35] \times [5, 20] \times [10, 25] \subset \mathbb{R}^4,  \quad (r_{\text{rib}}, r_{\text{top}}, p_{\text{rib}}, d_{\text{rib}}) \in \mathcal{D}
\end{equation}

\begin{figure}[ht]
    \centering
    \includegraphics[width=0.8\linewidth]{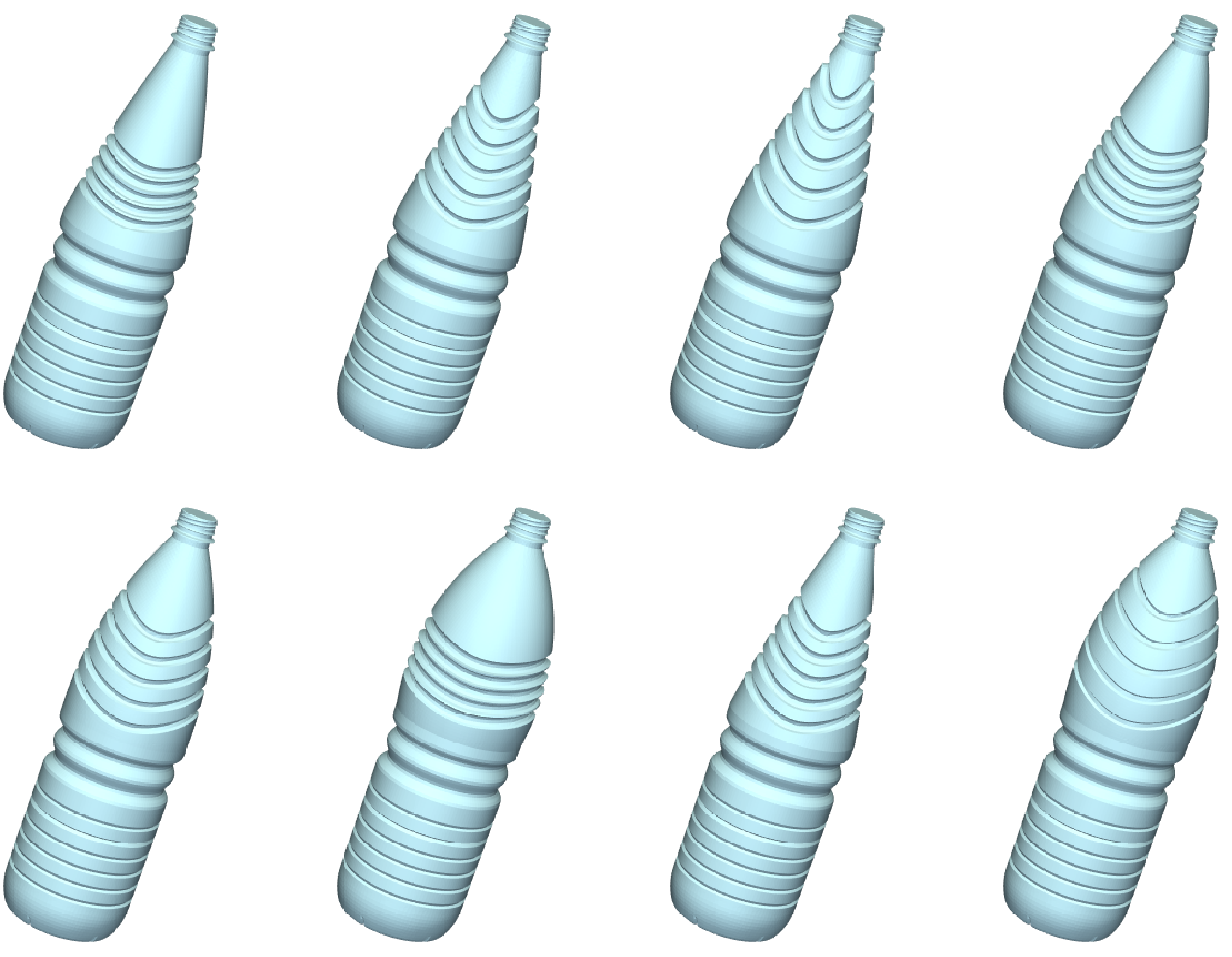}
    \caption{Representative samples of bottle geometries generated for this study. Top row represents two-parameter bottle samples while the bottom row shows four-parameter bottle samples. Four-parameter bottle family includes higher geometric variability due to a larger design space $\mathcal{D}$ and poses a more challenging setup for training a neural surrogate.}
    \label{fig:bottle_geoms}
\end{figure}

\begin{figure}[h]
    \centering
    \includegraphics[width=1\linewidth]{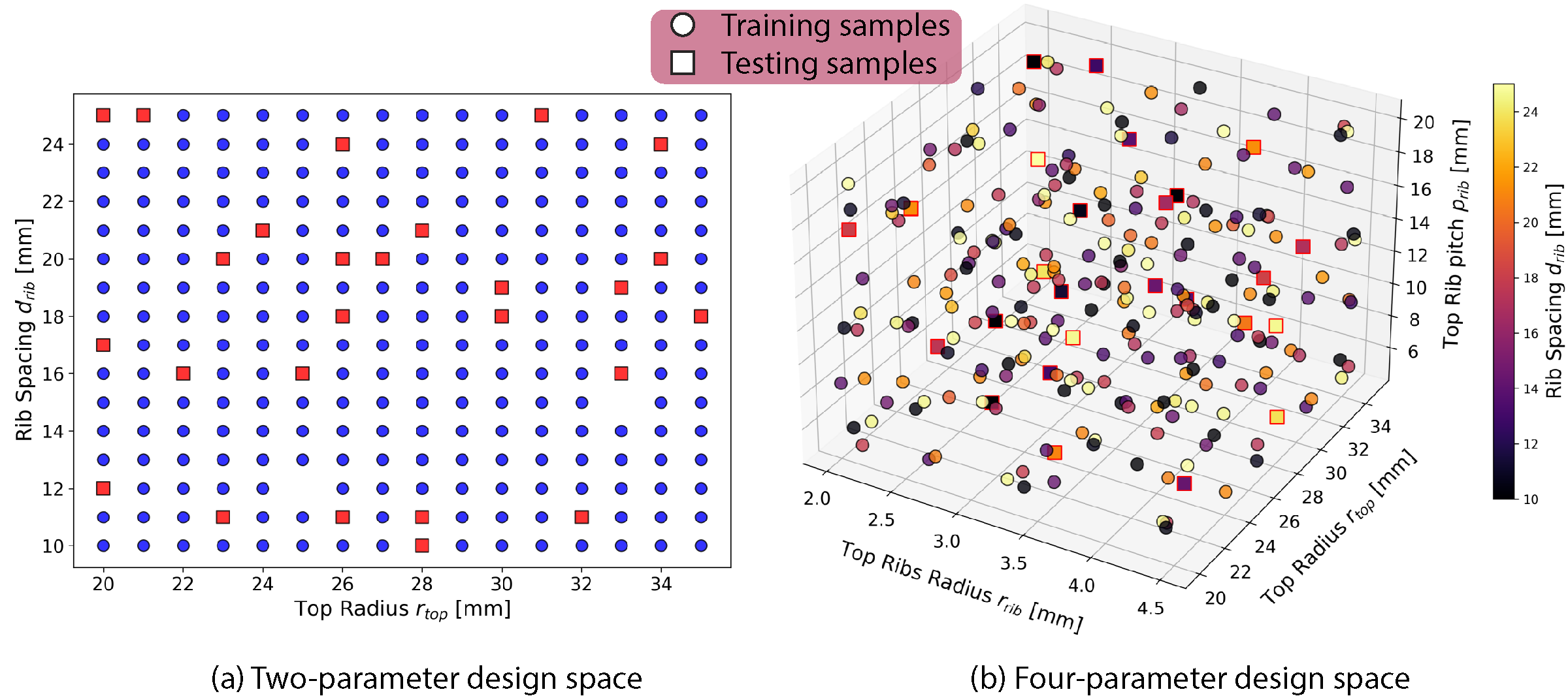}
    \caption{Visualization of design space $\mathcal{D}$ for the two bottle families. Here, circles denote the training samples while squares denote the test samples chosen in this study. (a) The two parameter design uses a uniform sampling strategy to cover the design space $\mathcal{D} \subset \mathbb{R}^2$. The missing samples in the design matrix are a result of bottle geometries that failed to converge in simulation. (b) The four-parameter design uses Adaptive sampling to efficiently cover the design space $\mathcal{D} \subset \mathbb{R}^4$ while minimizing computational cost. Note that the four-parameter design space results in more pockets of under-sampled regions.}
    \label{fig:design_space}
\end{figure}

The ranges of these parameters were selected to ensure adequate coverage of design space $\mathcal{D}$, and sample selection was performed using an Adaptive Design of Experiment strategy \cite{adaptiveDOE_Dassault} from the Process Composer App in Dassault Syst\`emes' \textbf{3D}EXPERIENCE platform. This ensures that the chosen designs efficiently span the design space and include sufficient variability for training a useful neural surrogate. For each case, a total of 254 unique designs are generated for FEA analysis. In  figure \ref{fig:bottle_geoms} we show representative samples from the two bottle families used in this study while figure \ref{fig:design_space} shows the location of chosen samples in the design space for the two bottle families. Note that these design parameters are only used for data generation and are not a part of our neural surrogate framework.

\subsection{Simulation details}
\label{subsec:data_gen}
We consider a displacement-driven problem to simulate top load condition for our bottle designs. In the context of bottle buckling analysis, prescribing displacement ensures that the simulation always produces a complete load–displacement curve. This includes capturing the peak reaction force, which represents maximum load-bearing capacity, as well as the potential subsequent softening phase that indicates the onset of buckling. To simulate, we first discretize the CAD designs into mesh-based representation within Abaqus, with meshes containing $\approx 20,000$ nodes and $\approx 20,000$ elements. The top cap of each bottle is prescribed a fixed downward displacement of 10 mm while the bottom is fully constrained. The bottle is modeled as an elastic-plastic material and the material, boundary condition, and loading conditions are kept same for all designs. Reaction forces were recorded at a reference point coupled to the nodes of the bottle top. The simulation was performed using Abaqus implicit solver \cite{Abaqus_implicit} which generates a history of nodal displacement field ($u_x, u_y, u_z$) and reaction force ($F_R$). The reaction force curve is analyzed to evaluate the resistance limit under top load and to determine the onset of buckling, which is indicated by the peak and subsequent decline in the reaction force history. The computational time for each simulation is $\approx 3$ minutes on one CPU. The simulation failed to converge for two cases in the two-parameter design family (see figure \ref{fig:design_space}) and therefore, we removed these samples from this study.

\section{Method}
\label{sec:method}
In order to create a \emph{geometry-aware} neural surrogate to learn the nodal displacements and time-dependent reaction forces, we propose a synergistic DeepONet-Transolver framework for learning both simulation outputs simultaneously. Before introducing details for our proposed approach, we provide a brief background on Transolver and DeepONet in the following sections.

\subsection{Transolver}
Conventional neural surrogates designed for mapping solutions to PDEs on multiple geometries rely on propagation of node coordinates and/or the node connectivity information. This approach enables learning low-level relationships between mesh points but not capturing the  
correlation between regions of geometry under similar physical states. The objective function for such surrogates is to map mesh-based information to the nodal outputs obtained from solving the PDE. Transolver solves this by shifting perspective: instead of directly working with raw mesh information, it groups them into a smaller set of latent physical states that exhibit similar response. This allows mapping of similar regions in the domain to physically-consistent slices of data aligned with the PDE solution in these regions. 

We consider input as a mesh-based geometric representation $\mathbf{g} \in \mathbb{R}^{N \times C}$, where $N$ represents the number of nodal points and $C$ represents the input feature of each nodal point. These input features may include spatial coordinates, normal vectors for each point, or zero-level set representations such as SDF. The set of input points can be mapped to $S$ physically-consistent slices using the following steps:
\begin{enumerate}
    \item Project $\mathbf{g}$ to a higher dimension, $\mathbf{v}$, using a linear projection such that
    \begin{equation}
        \phi: \mathbb{R}^{N \times C} \longrightarrow \mathbb{R}^{N \times D}, 
    \end{equation}
    where $\phi$ represents a multi-layer perceptron with activation function and $\mathbf{x} = \phi(\mathbf{g})$.
    \item Rearrange the projected feature vector $\mathbf{x}$ into $S$ slices across $H$ attention heads
    \begin{equation}
        \mathbf{x} \xrightarrow[]{\text{Rearrange}} \mathbf{x}^{*}, \quad \mathbf{x}^{*} \in \mathbb{R}^{H \times N \times S}. 
    \end{equation}
    \item Attribute weight $\mathbf{w}$ of each mesh point $N$ to each slice using softmax operation along slice dimension $S$
    \begin{equation}
        \mathbf{w}_{hjs} = \dfrac{e^{\mathbf{x}^{*}_{hjs}}}{\sum \limits_{s=1}^{S} e^{\mathbf{x}^{*}_{hjs}}}, \quad \mathbf{w}_{hjs} \in \mathbb{R}^{H \times N \times S}. 
    \end{equation}
\end{enumerate}
The token embeddings for dot product attention are obtained for each slice by aggregating across all $N$ points
    \begin{equation}
        \mathbf{z} = \dfrac{\sum\limits_{j=1}^{N} \mathbf{w}_{hjs} \mathbf{\tilde x^{*}}}{\sum \limits_{j=1}^{N} \mathbf{w}_{hjs}}, \quad \mathbf{z} \in \mathbb{R}^{H \times S \times D_{h}}, 
    \end{equation}
    where $\mathbf{\tilde x^{*}} \in \mathbb{R}^{H \times N \times D_{h}}$ represents linear projection of the mesh representation $\mathbf{g}$ using another set of MLP, $\psi : \mathbb{R}^{N \times C} \longrightarrow \mathbb{R}^{N \times (H \times D_{h})}$, $D_{h}$ denotes the width of the last MLP layer. These token embeddings $\mathbf{z}$ represent the weighted contribution to each slice from nodal points with similar physical features.

    Using these physically-consistent tokens, we apply the dot product attention mechanism to obtain the attention values
    \begin{equation}
        \mathbf{z}' = Softmax\left( \dfrac{\mathbf{q} \cdot \mathbf{k^{T}}}{\sqrt{D_h}}\right) \mathbf{v}, \quad \mathbf{q, k, v} \in \mathbb{R}^{H \times S \times D_{h}}, \mathbf{z}'\in \mathbb{R}^{H \times S \times D_{h}}, 
    \end{equation}
    where $\mathbf{q, k, v} = \textit{Linear}(\mathbf{z})$ represent the linear projection of token embeddings $\mathbf{z}$ using a linear layer. During training, the token embeddings in each slice are updated to align themselves according to the similarity of output feature values due to the dot product. Points showing similar output values will be aligned to minimize the dot product during back propagation. This is similar to the token alignment scheme of word embeddings in large language models where the positional proximity of words helps to update the tokens during back propagation.

    For reconstructing the required output field $\mathbf{u} \in \mathbb{R}^{N \times C'}$, de-slicing operation is performed using the slice weights $\mathbf{w}_{hjk}$ combined with the attention values using summation across slice dimension $S$ to obtain representation $\mathbf{y}$ given by
    \begin{equation}
        \mathbf{y} = \sum \limits_{s=1}^{S} \mathbf{z}'_{hsd} \mathbf{w}_{hjs}, \quad \mathbf{y}\in \mathbb{R}^{H \times N \times D}.
    \end{equation}
    This is reshaped into feature embedding $\mathbf{y}^{*}$ across all mesh points $N$ through
        \begin{equation}
            \label{eqn:transolver_embedding}
            \mathbf{y} \xrightarrow[]{\text{Rearrange}} \mathbf{y}^{*}, \quad \mathbf{y}^{*} \in \mathbb{R}^{N \times D}.
        \end{equation}
    Finally, the output features $\mathbf{u}$ are obtained following translations through MLPs
    \begin{equation}
        \mathbf{u} = \phi(\mathbf{y}^{*}), \quad \mathbf{u} \in \mathbb{R}^{N \times C'}, 
    \end{equation}
    where $C'$ represents the number of output features of the solution. 
\subsection{DeepONet}
For Banach spaces $\mathcal{U}$ and $\mathcal{S}$ defined as
\begin{align}
    &\mathcal{U} = \{\Omega; u: \mathcal{X} \to \mathbb{R}^{d_u}\}, \quad \mathcal{X}\subseteq \mathbb{R}^{d_x}\\
    &\mathcal{S} = \{\Omega; s: \mathcal{Y} \to \mathbb{R}^{d_s}\}, \quad \mathcal{Y}\subseteq \mathbb{R}^{d_y}, 
\end{align}
where $\Omega \subset \mathbb{R}^D$ denotes the open bounded domain, $\mathcal{U}$ represents the set of admissible input functions and $\mathcal{S} $ corresponds to the output function space. The neural operator approximation that maps $\mathcal{U} \longrightarrow \mathcal{S}$ is defined as
\begin{equation}
    \mathcal{H}_{\boldsymbol{\theta}}: \mathcal{U} \longrightarrow \mathcal{S}, \quad \boldsymbol{\theta} \in \mathbf{\Theta}, 
\end{equation}
where $\mathbf{\Theta}$ denotes the finite dimensional parameter space of the neural surrogate. In practice, the model is trained on a dataset of paired input-output functions $\mathbf{M} = \left\{(u^{(i)}, s^{(i)})\right\}_{i=1}^n$. The deep operator network (DeepONet) \cite{deeponet_lulu_2021}, introduced as a neural framework inspired by the universal operator approximation theorem \cite{universal_approx}, has emerged as a powerful tool in operator learning for PDE tasks. Its design couples two neural networks: a branch network, which encodes the input function $\mathbf{u}$ sampled at finite sensor locations $\{x_1, x_2, ..., x_m \}$ and a trunk network, which processes the spatial-temporal coordinates $\zeta \in \mathbb{R}^{d}, \zeta := (x,y,z,t)$ at which the operator is evaluated. 

For a given realization of input function $u_1$, the DeepONet representation of the operator output at $\zeta$ is
\begin{equation}\label{eq:output_deeponets}
      \mathcal H_{\boldsymbol \theta}(u_1)(\zeta) = \sum_{i = 1}^p b_i \cdot tr_i = \sum_{i = 1}^{p}b_i(u_{1}(x_1), u_{1}(x_2), \ldots, u_{1}(x_m))\cdot tr_i(\zeta),
\end{equation}
where $\{b_{i}\}_{i=1}^p$ are output embeddings from the branch network and $\{tr_{i}\}_{i=1}^p$ are embeddings from the trunk network. The network parameters $\boldsymbol{\theta} = \left(\mathbf{W}, \mathbf{b} \right)$ are obtained by minimizing a cost function generally through supervised learning. 

\subsection{Hybrid DeepONet-Transolver framework}
\begin{figure}[h]
    \centering
    \includegraphics[width=1\linewidth]{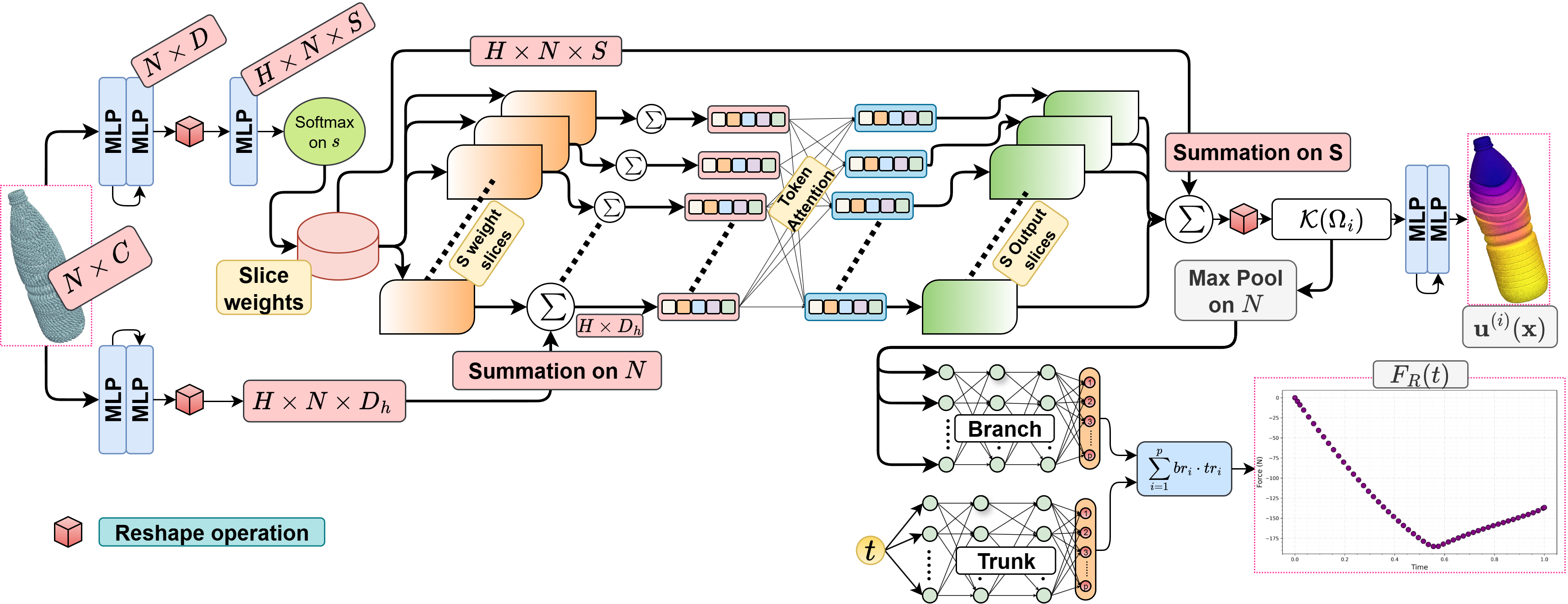}
    \caption{DeepONet-Transolver framework for simultaneous prediction of nodal displacements $\mathbf{u}(\mathbf{x})\in \mathbb{R}^{N \times 3}$ and time-dependent reaction forces $F_{R} (t)$ for the top load compression analysis of packaging bottles. Here, $N$ represents the number of nodal points that defines a point cloud for geometry $\Omega_{i}$, and $C$ represents the input feature dimension,  which comprises of the coordinate locations and surface normals. $D$ represents the width of the MLP layer, $H$ represents the number of attention heads for self-attention, $D_{h}$ represents the attention head dimension, and $S$ represents the number of slices into which the inputs features are distributed based based on their proximity. The Transolver network predicts the nodal displacement $\mathbf{u}(\mathbf{x})$ at the last time step $t=1$ in the simulation while the integrated DeepONet is used for predicting the reaction force using feature embeddings learnt during Transolver training as input to its Branch network. Both networks are trained concurrently.} 
    \label{fig:TransDON_Schematic}
\end{figure}

The Transolver framework relies on output function values at nodal points to learn and align token embeddings. Surrogate modeling of bottle buckling includes mapping the nodal displacements at the end of simulation time as well as the reaction force experienced at the top of the bottle during this process. While the displacements at the end of simulation can be treated as a static mapping between geometric features to output function and can be mapped using Transolver, the time-dependent reaction force requires additional treatment. In this study, we propose an integrated Transolver and DeepONet approach for simultaneously mapping both the static displacements $\mathbf{u}(\mathbf{x}, t = t_{end})$ as well as the reaction force $F_{R}(t)$. Figure \ref{fig:TransDON_Schematic} shows a schematic of the integrated framework proposed in this study. Here, the Transolver framework is used for learning the nodal displacement vector for the last simulation time step. The feature embedding $\mathbf{y}^{*} \in \mathbb{R}^{N \times D}$ from equation \ref{eqn:transolver_embedding} is obtained as a result of mapping 
\begin{equation} \label{eqn:y-star}
    \mathcal{K}: \mathbb{R}^{N \times C} \longrightarrow \mathbb{R}^{N \times D}, \quad \mathbf{y}^{*} = \mathcal{K}(\Omega_i), 
\end{equation}
where $\mathbf{y}^{*}$ represents a latent representation, and we use this as an input to DeepONet's branch network after applying a max pooling operation across nodal points $N$. This helps to  create a geometry-specific function representation for individual geometries $\Omega_{i}$ during its mapping to displacements $\mathbf{u^{(i)}(\mathbf{x})}$. The trunk network uses as input the time steps at which reaction force ($F_R$) is measured during the simulation and is trained with data to learn the map according to equation \ref{eq:output_deeponets}
\begin{equation} \label{eqn:deeponet_mapping}
    \mathcal{H}_{\mathbb{\theta}}: Y^{*} \times T \longrightarrow F, \quad Y^{*} \subseteq \mathbb{R}^{D}, \quad T \subseteq \mathbb{R}, \quad F \subseteq \mathbb{R}.
\end{equation}

\section{Experiments}
We evaluate our hybrid DeepONet-Transolver framework with two sets of bottle data: two-parameter design and four-parameter designs. Bottle shapes generated by varying two parameters will have more similarity as compared to four-parameter designs and this enables us to evaluate the performance of our framework for \emph{geometry aware} tasks with different levels of geometric variations. In this section, we present the network architectures  and training details along with results of our experiments with the two data sets.
\label{sec:experiments}
\subsection{Implementation details}
The data available for our experiments contains nodal displacements at the end of simulation, reaction force evolution in time, and the mesh-based representation of the bottle geometries. For network training, we generate an input-output dataset $\mathbf{D}: \{(X_{i}, Y_{i})\}_{i=1}^{N_{g}}$ with $$ X_{i} := (\Omega_i, t), Y_{i} := (\mathbf{u}^{(i)}(\mathbf{x}), R_{F}^{(i)}(t)),$$where $N_g$ denotes the number of geometries. Each geometry input $\Omega_{i} := \{\mathbf{x}_{j}, \mathbf{n}_{j}\}_{j=1}^{N}$ in our study is represented by set of $N \simeq 20,000$ mesh nodes and their corresponding normal vectors ($\mathbf{n}$) extracted using the surface mesh. The output $\{\mathbf{u}(\mathbf{x}_{j})\}_{j=1}^{N}$ represents the nodal displacements for each $\Omega_i$ at the end of the simulation step $t = t_{end}$ while $F_R(t)$ represents the reaction force experienced at each simulation time step. 

\subsection*{Data pre-processing}
We randomly divide the 254 data samples into training (228 samples) and testing (26 samples) using a 90-10\% data split. Prior experiments with a 80-20\% split resulted in lower overall accuracy ($\sim 2-3 \% \,\text{higher error}$) and we note that a larger training sample size is important to encompass the design space effectively.  To improve training stability, both input and output data are normalized using the empirical mean ($\mu$) and standard deviation ($\sigma$) of training data along individual dimensions. For a given variable $z$, the normalized form, $\tilde z$ is defined as 
\begin{equation}
    \tilde z = \dfrac{z - \mu}{\sigma + \epsilon}, \quad \epsilon << 1, 
\end{equation}
where $\epsilon$ is a small positive constant ($1e^{-8})$ added to denominator for numerical stability. The geometry data $\Omega_{i} := \{\mathbf{x},  \mathbf{n}\} \in \mathbb{R}^{6}$ comprises of the nodal coordinates $\{x, y, z\}$ and the normal vectors ($n_x, n_y, n_z$) concatenated together. The normal vectors are obtained from individual bottle surface meshes using the PyVista \cite{pyvista} library. The output displacement $\mathbf{u}(\mathbf{x}) := \{u_x, u_y, u_z\}$ contains the displacements along the three axes. The reaction force, $F_R(t) \in \mathbb{R}$ is evaluated at each time step of simulation, where $t \in [0, 1]$.

\subsection*{Network setup and training}
For Transolver, we use an architecture similar to the original work with $H = 8$ attention heads, MLP layers with 128 hidden features, four physics-based attention blocks with 32 slices, and GELU activation \cite{gelu}. Layer normalizations, and residual connections are applied as per the original work. We use Pytorch's Distributed Framework for data parallelism across eight NVIDIA RTX A5500 GPUs to improve the training speed. The Transolver framework is trained to predict the three displacement vectors $u_x, u_y, u_z$. For the DeepONet module, we use a branch network with three hidden layers with a width of 128 with GELU activation. Prior to using the Transolver's encoded output $\mathbf{y}^{*}$ with the branch, we also apply a layer normalization for training stability. The trunk network comprises of three hidden layer with width of 128 neurons each and also contains residual connection between layers. The branch and trunk networks generate outputs $\mathbf{y}_{br} \in \mathbb{R}^{1 \times 128}$ and $\mathbf{y}_{tr} \in \mathbb{R}^{N_t \times 128}$ respectively, which after summation as per equation \ref{eq:output_deeponets} yields the required output reaction outputs $F_R$ measured at $N_t$ time steps.

We train the network by minimizing the mean $L^1$ error between predicted outputs and ground truth values. This choice of loss function was heuristically found to yield better prediction results as compared to standard mean squared loss. The optimization objective is given by
\begin{equation} \label{eqn:loss_def}
    \mathcal{L} = \Vert \hat{u} - u \Vert_{1}, \quad \hat{u}, u \in \mathbb{R}^{N \times 1}
\end{equation}
where $\hat{u}$ represents the output from the neural surrogate and $N$ denotes the number of nodal points defining each geometry. We calculate the loss term for each displacement component and reaction forces separately and obtain the complete loss term given as
\begin{equation} \label{eqn:total_loss}
    \mathcal{L}_{total} = \mathcal{L}_{u_x} + \mathcal{L}_{u_y} + \mathcal{L}_{u_z} + \mathcal{L}_{F_R}.
\end{equation}
The network parameters for Transolver and DeepONet are optimized simultaneously using Adam optimizer with an initial learning rate of $1e^{-3}$. The learning rate is dynamically adjusted during training using OneCycleLR scheduler \cite{onecycleLR}, which increases the learning rate to a maximum and then gradually decreases it, improving convergence stability and final performance. The network is trained for 10000 epochs with a batch size of one with evaluation step once every 50 epochs. To assess inference quality, we use the relative $L^2$ norm of error averaged over the number of test samples $N_{test}$ across $N$ nodal points in each sample
\begin{equation} \label{eqn:L2_err}
    Err = \dfrac{1}{N_{test}}\sum \limits_{i=1}^{N_{test}}\dfrac{\Vert \hat{u}^{(i)} - u^{(i)} \Vert _{2}}{\Vert u^{(i)} \Vert_{2}}, \quad \hat{u}, u \in \mathbb{R}^{N \times 1}.
\end{equation}

\subsection{Main results}
In this section, we summarize our experimental findings and provide analysis of our insights from this study.
\subsection*{Error estimates}
\label{subsec:main_Results}
In table \ref{tab:results_err_summary}, we present a summary of the model's accuracy across both sets of bottle designs. We include error metrics for the mean error observed on the test data with equation \ref{eqn:L2_err} and also provide the error estimates of the median, lowest and highest error values we observe across individual samples. For the two-parameter bottle designs, the average $Err$ for the component-wise displacement predictions across all 26 test cases varies between $\approx 1\% - 13\%$ while the mean error for the reaction force was $\approx 0.30\%$. A lower median value of error across all three displacements as compared to mean errors suggest that the model prediction for larger proportion of test cases lies within the $0.6-7.7\%$ error range. Additionally, one example in our test case exhibits significantly higher error for $u_y$ as noted by the max error value in table \ref{tab:results_err_summary}.

\begin{small}
\begin{table}[ht] 
\centering
\caption{Summary of $L^2$ relative errors obtained for the two different bottle families. While the overall errors for test cases are reasonable, few test cases show high errors due to these particular samples being under-represented in the training dataset. Note that the best and worst case error estimates represent error across different data samples.}
\resizebox{0.9\linewidth}{!}{%
\begin{tblr}{
  column{even} = {r},
  column{3} = {r},
  column{5} = {r},
  column{7} = {r},
  column{9} = {r},
  cell{1}{2} = {c=4}{c},
  cell{1}{6} = {c=4}{c},
  cell{2}{2} = {c},
  cell{2}{3} = {c},
  cell{2}{4} = {c},
  cell{2}{5} = {c},
  cell{2}{6} = {c},
  cell{2}{7} = {c},
  cell{2}{8} = {c},
  cell{2}{9} = {c},
  vline{3} = {1}{0.05em, solid},
  vline{6} = {1-6}{1pt, dashed},
  hline{1,7} = {-}{0.08em},
  hline{2} = {2-9}{0.08em},
  hline{3} = {-}{0.05em},
}
 & $Err$: 2-parameter &  &  &  & $Err$: 4-parameter &  &  & \\
 & $u_x$ & $u_y$ & $u_z$ & $F_R$ & $u_x$ & $u_y$ & $u_z$ & $F_R$\\
Mean & 0.1049 & 0.1288 & 0.0088 & 0.0030 & 0.1107 & 0.1309 & 0.0245 & 0.0240\\
Min & 0.0476 & 0.0424 & 0.0055 & 0.0003 & 0.0336 & 0.0380 & 0.0081 & 0.0044\\
Max & 0.2654 & 0.6616 & 0.0267 & 0.0093 & 0.3400 & 0.7090 & 0.0803 & 0.0777 \\
Median & 0.0770 & 0.0616 & 0.0065 & 0.0025 & 0.1013 & 0.1084 & 0.0159 & 0.0166
\end{tblr}
}
\label{tab:results_err_summary}
\end{table}
\end{small}

The mean error values observed for the four-parameter designs are slightly higher as compared to the two-parameter bottle design,  which is expected since the four-parameter bottle shapes has a larger design space, $\mathcal{D} \in \mathbb{R}^4$. This means that the design samples generated in this design space have larger separation between them, thus leading to a sparser-dataset as compared to the two-parameter samples. Consequently, the entire design space is under-represented in this data. Thus, higher mean and median errors are observed for this data set ranging between $\approx 2.5-13\%$ for the displacements and $\approx 2.4\%$ for the reaction force. A significantly higher error for $u_y$ for one particular sample is also observed in this dataset leading to a higher mean error as compared to the median error value.  Upon closer inspection, we found that the displacement distribution for this case deviated markedly from the distributions represented in the training set, which can lead to poor generalization. Since the neural surrogate had not been trained on a response pattern of this type, prediction on this particular sample is essentially in an extrapolation regime.

\subsection*{Point-wise errors for displacement}
\begin{figure}[t]
    \centering
    \includegraphics[width=1\linewidth]{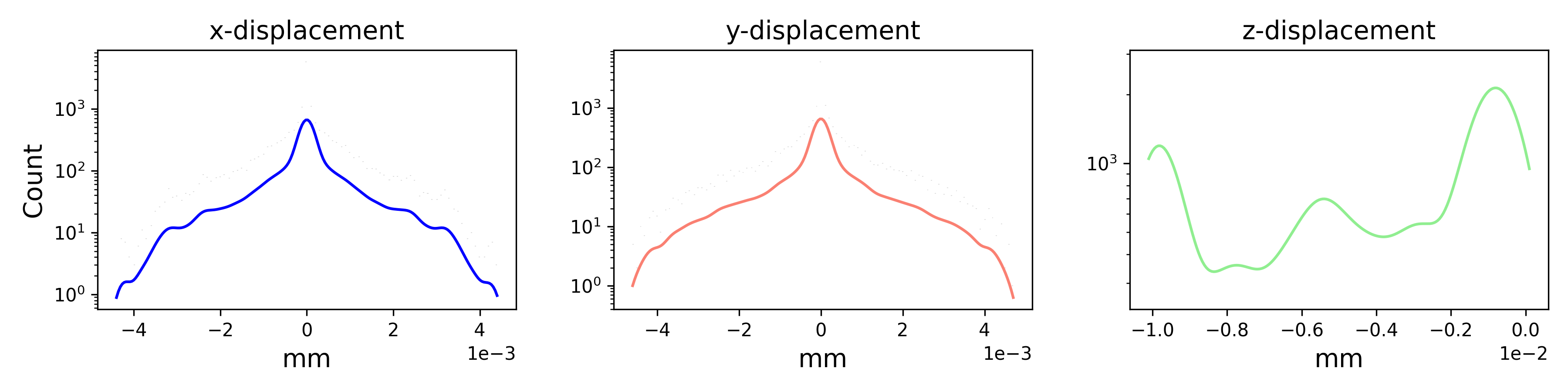}
    \caption{Plot showing distribution of displacement values for a data sample. A log scale is used for the vertical axis to ensure data is scaled appropriately to show the peak of data frequency around displacement values of 0.0 for $u_x$ and $u_y$. Such unbalanced distribution can bias the average $L^2$ error estimates.}
    \label{fig:distribution_displacements}
\end{figure}

While $L^2$ error norms are useful for estimating the overall model effectiveness, an assessment of point-wise errors is important here since the displacement data contains a significant proportion of nodes with zero displacement. In figure \ref{fig:distribution_displacements} we show a distribution of displacement data for a representative sample, where a significantly high data accumulation around zero can be seen. This happens due to the fixed boundary condition that exists at the bottom of the bottle, which leads to the ribbed region exhibiting highest displacements during top load. The average $L^2$ error in such scenarios will be dominated by how well the model can predict those zeros, potentially overshadowing the performance on the smaller but more important nonzero regions.

\begin{figure}[h]
    \centering
    \includegraphics[width=1\linewidth]{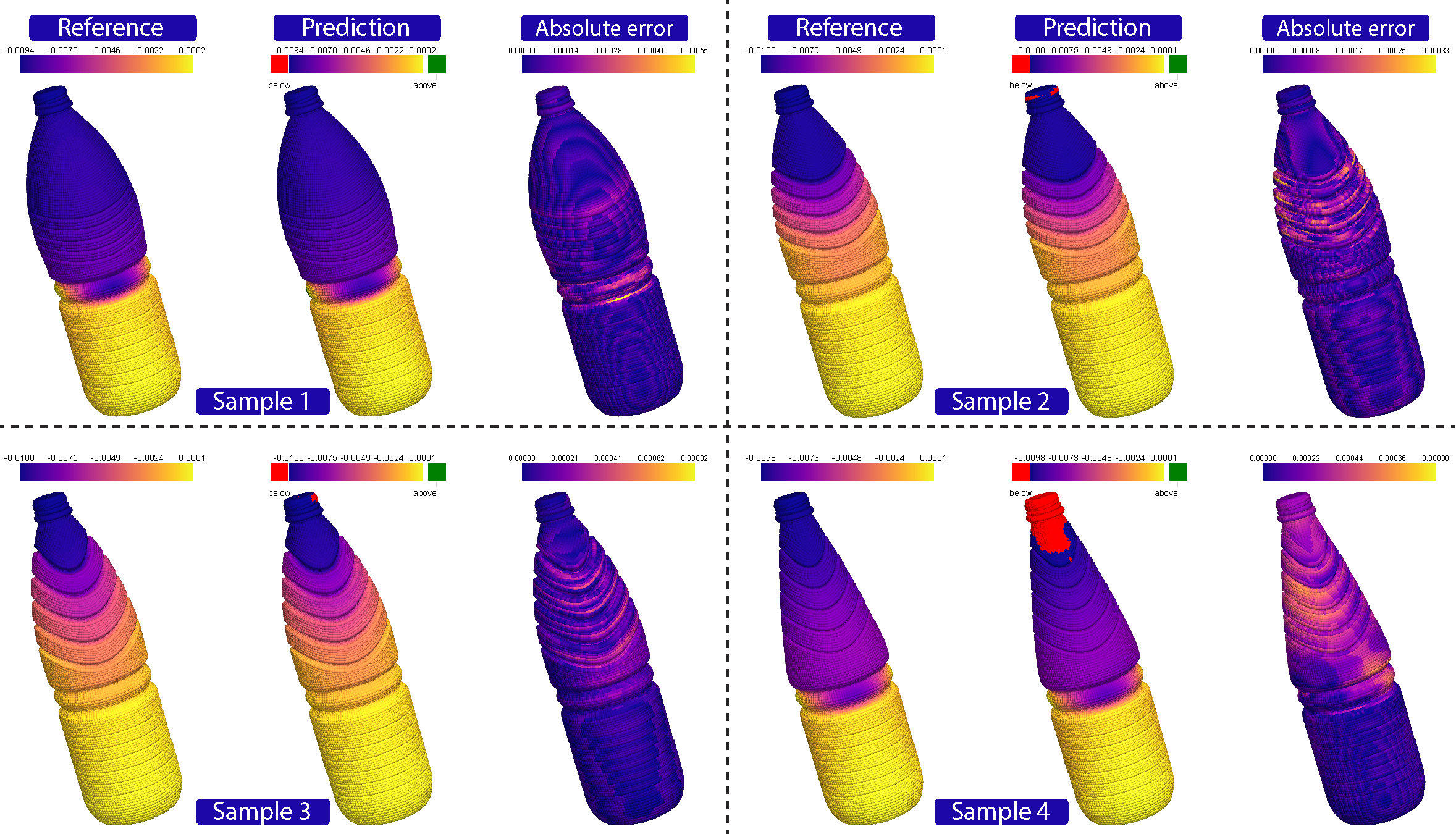}
    \caption{Point-wise absolute errors in displacement $u_z$ for four-parameter bottle family test samples. Points in red indicate model prediction that is below the minimum displacement value of reference solution while the points in green indicate predictions that are larger than the maximum value of reference solution. }
    \label{fig:4param_predictions}
\end{figure}

Figure \ref{fig:4param_predictions} presents point-wise $u_z$ displacement predictions for four test samples of the four-parameter bottle family. Across all samples, we observe that the predicted displacement field closely matches the reference solution, capturing both global deformation pattern and localized variations. This indicates that the model can successfully predict the underlying deformation behavior across different geometric shapes with good accuracy. The absolute error plot reveals that the large proportion of the point-wise errors are small (of order $10^{-4} - 10^{-3}$) and are concentrated in specific regions rather than uniformly spread across the domain. The high-error regions occur primarily near the bottle neck and cap region and along transition regions, where corrugated rib patterns meet the smooth surface. Also, in some sample predictions, isolated regions (indicated in red) are found where the model under-predicts relative to the reference solution. The magnitude of absolute error shows that although small localized discrepancies exist, the model predictions are quantitatively accurate overall. Also, the prediction accuracy is consistent across all test samples (except an anomaly with very high error for $u_y$ displacement), which suggests that the model generalizes well across different geometries. These findings are consistent across the two-parameter bottle designs as well.

\begin{figure}[t!]
    \centering
    \begin{subfigure}[t]{1\textwidth}
        \centering
        \includegraphics[width=1\linewidth]{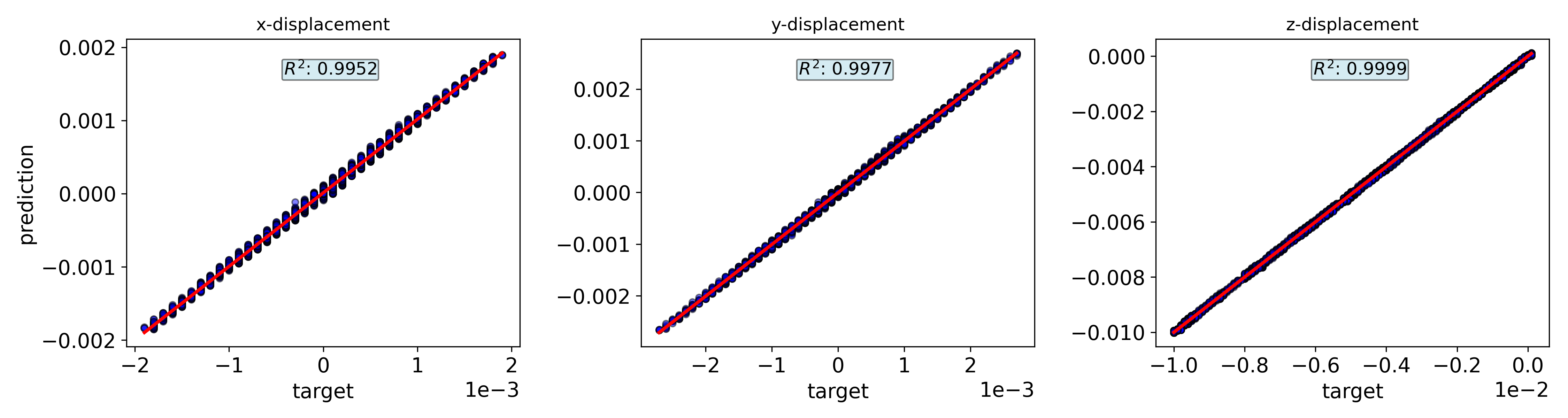}
        \caption{Scatter plot comparing model prediction and reference displacement solutions for a two-parameter geometry sample with $Err\{u_x, u_y, u_z\} = \{0.0873, 0.0465, 0.0078\}$.}
    \end{subfigure}%
    \vfill
    \begin{subfigure}[t]{1\textwidth}
        \centering
        \includegraphics[width=1\linewidth]{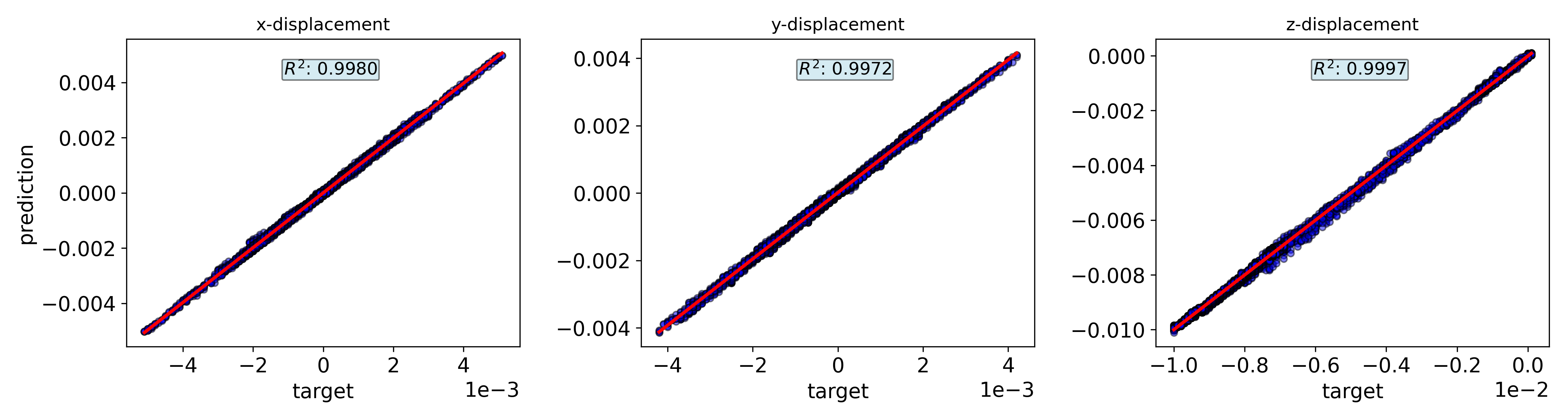}
        \caption{Scatter plot comparing model prediction and reference displacement solutions for a four-parameter geometry sample with $Err\{u_x, u_y, u_z\} = \{0.0417, 0.0442, 0.0100\}$}
    \end{subfigure}
    \caption{Scatter plot showing goodness-of-fit of the model for the two bottle design class. The plots show a strong linear correlation between the model's predicted displacements and the target values for all three spatial components of displacement.}
    \label{fig:R2_plot}
\end{figure}

To quantify the predictive performance of the proposed learning framework across multiple geometrical configurations, we evaluate the coefficient of determination ($R^2$) between model predictions and reference solutions for the three displacement components. Figure \ref{fig:R2_plot} shows scatter plots for a test sample from the two classes of bottle designs investigated in this study. The resulting $R^2$ values for the test samples exceed values of $0.99$ in most cases across all displacement components, indicating that the learned surrogate captures a significant part of the variance in target fields.  

\subsection*{Reaction force predictions}
\begin{figure}[h]
    \centering
    \includegraphics[width=1\linewidth]{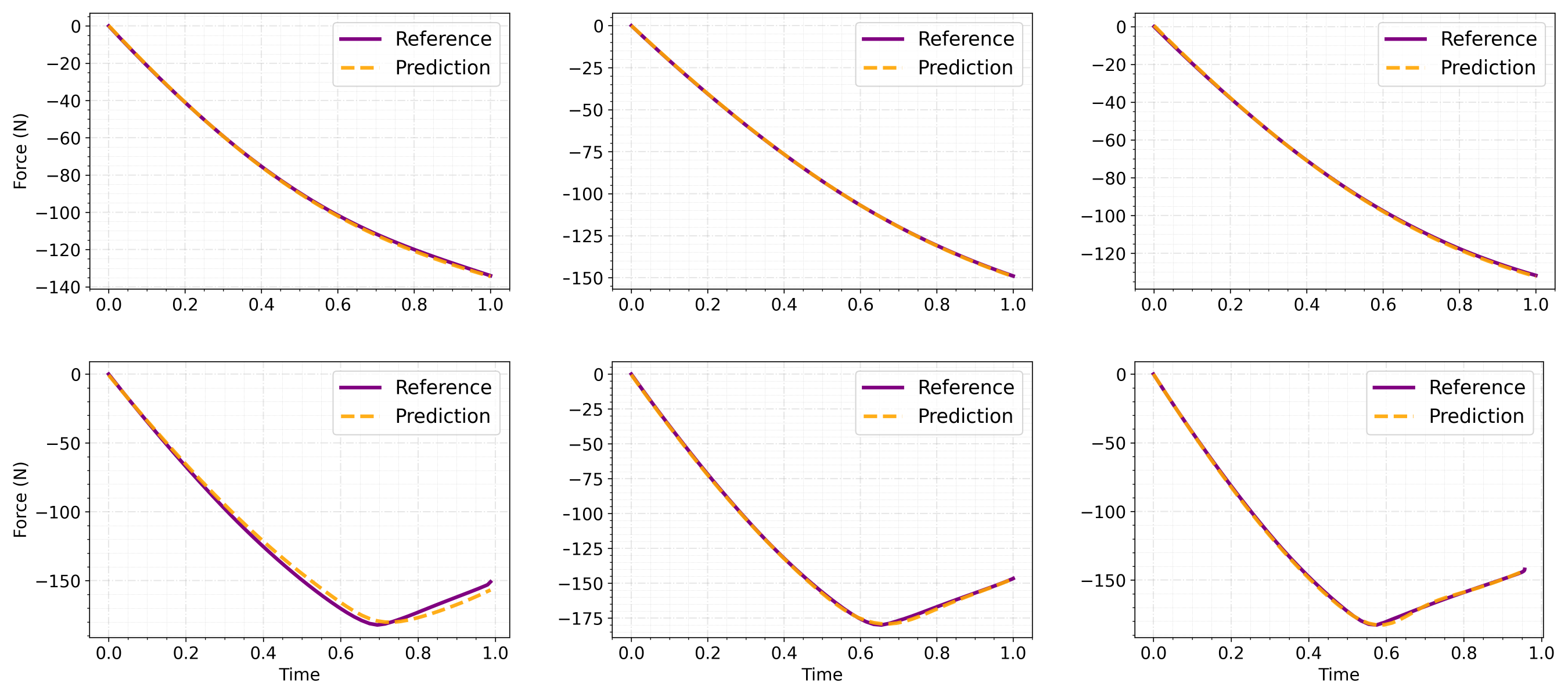}
    \caption{Comparison of predicted and reference reaction forces versus time for two-parameter (top-row) and four-parameter (bottom-row) bottle designs. The plots show a good agreement between our model's predictions (orange) and the ground truth data(purple). The model accurately captures the mechanical response across all designs, including the complex nonlinear behavior and buckling events evident in the four-parameter designs.}
    \label{fig:reaction_force_pred}
\end{figure}

Predicting the magnitude of reaction force as it evolves in time is a key element in understanding the buckling characteristic of different bottle designs. In our framework, the integrated DeepONet predicts the reaction force as a function of time using latent features learnt during Transolver training. Figure \ref{fig:reaction_force_pred} compares the predicted reaction forces against reference data obtained from numerical simulations  across different data samples for both sets of data. The plot demonstrates a high degree of accuracy, especially for the samples representing the two-parameter bottle geometries (top row). Note that the profile of the reaction force is relatively simple and monotonic for the two-parameter bottle designs because these geometries do not exhibit buckling behavior during simulation. On the other hand, the reaction force profile for the four-parameter geometry samples (bottom row) exhibits complex, nonlinear behavior including a gradient-reversal point which signifies the onset of buckling. Beyond this buckling point, the reaction force experienced starts to reduce. A reasonable match to reference solutions indicated by this plot highlights the model's ability to capture significant physical events across multiple geometric designs accurately. Here, DeepONet offers a key advantage in time-dependent prediction due to its resolution invariant basis functions generated by the trunk network. This allows us to use as input the time steps from numerical solver and thereby eliminate the need for extrapolation, which would be needed in certain samples where the simulation terminates early ($t_{end} < 1$). 

\section{Limitations and Challenges} 
\label{sec:lim_future_work}
In this work, we demonstrated the ability to capture complex nonlinear responses such as buckling events with high accuracy across complex geometric designs using a hybrid DeepONet-Transolver framework. This highlights the promise of geometry-aware neural surrogates as efficient alternatives to computationally intensive finite element simulations in computational mechanics. Nevertheless, we note here some key limitations and challenges observed during this study. One important challenge relates to the sensitivity of surrogate prediction to the choice of training samples, particularly as the design space expands. With a sparse set of training samples available, it becomes crucial to ensure that these samples adequately represent the variability of the design space. Selecting training cases from sparser or more diverse regions of the design space can improve the effectiveness of such surrogates. However, this strategy is less feasible in non-parametric settings, where the design space is unknown or difficult to characterize and often, random sample selection is used in such cases. 

Additionally, the two sets of bottle geometry considered in this study are simplified parameterizations involving two and four design factors. While this problem setting may be sufficient to demonstrate the feasibility and accuracy of the proposed framework, it does not fully capture the complexity of the real-world industrial design problem, where bottle geometries may depend on seven or more parameters. Such higher-dimensional design spaces can introduce richer geometric variations and significantly increase the difficulty of achieving robust generalization. Our preliminary experiments with seven-parameter bottle designs with a smaller sample size revealed that the model's accuracy decreases notably under such conditions. Hence, extending this methodology to handle higher-dimensional parametric representations is necessary, and we continue to explore ways to improve the capability of current framework to more complex designs with smaller sample size to replicate industrial design tasks.

\section{Summary}   
\label{sec:summary}
We proposed a neural surrogate for learning solution maps for \emph{geometry aware} problems in scientific computing. Specifically, we created a hybrid framework using Transolver and DeepONet for predicting static nodal displacements and time-dependent reaction force simultaneously across varying PET bottle designs under top compressive loading. The dataset was constructed from families of bottle designs parameterized by two and four design variables. For each geometry, nodal displacement and reaction forces were computed using Abaqus. This dataset was used for training our framework that relies on Transolver's physics-based attention mechanism to predict nodal displacements and the DeepONet for predicting time-dependent reaction force experienced at the top of the bottle during compression. We evaluated our framework's performance using the $L^2$ relative norm as our error metric calculated independently across each output.
During inference, we observe good prediction accuracy across both the two- and four-parameter designs. For the two-parameter bottle design set, we observed a mean $L^2$ relative error between $1-14\%$ for the displacement fields ($u_x, u_y, u_z$) and $\approx 0.4\%$ for the reaction force ($F_R$). For the more challenging four-parameter designs, we observed slightly higher error values, which are still within acceptable range for this application. Point-wise prediction plots further confirmed that the learned surrogate can capture fine-scale variations of the solution fields with good fidelity. The point-wise absolute error value plot reveals that the large proportion of the point-wise errors are small (of order $10^{-4} - 10^{-3}$) and higher errors are concentrated in specific regions rather than uniformly spread across the domain. These high-error regions occur around the bottle neck and ribbed region of the geometry where strong 
nonlinear deformation exists during bottle buckling. We also evaluated the goodness-of-fit of our model by estimating the coefficient of determination $R^2$ between our model's displacement prediction and reference solution and for most test cases, and we found high $R^2$ values of $0.98$ and higher across all displacement fields. The accuracy of reaction force across the two- and four-parameter design sets was $0.39\%$ and $\approx 3\%$, which signifies that the integrated DeepONet module generates a reasonable approximation for time-dependent outputs using latent embeddings from Transolver. This approach extends the Transolver model's applicability to other problems requiring multi-task solutions.

Overall, this work establishes a promising direction for using deep learning models as efficient surrogates for multi-geometry problems in computational mechanics, enabling rapid evaluation of candidate designs without repeated high-fidelity simulations. Our future endeavors are focused on extending the use of this framework for problems with larger design space and improve computational efficiency.

\section*{Acknowledgments}
This research was conducted using computational resources and services at the Center for Computation and Visualization, Brown University.

\section*{Funding}
This study was funded by Dassault Syst\`emes. VK and GEK also acknowledge support from AFOSR Multidisciplinary Research Program of the University Research Initiative (MURI) grant FA9550-20-1-0358, ONR Vannevar Bush Faculty Fellowship (N00014-22-1-2795), and U.S. Department of Energy project SEA-CROGS (DE-SC0023191).

\section*{Author contributions}
\noindent Conceptualization: VK \\
Investigation: VK\\
Visualization: VK\\
Supervision: JB, CNG, VO, GEK \\
Writing—original draft: VK, JB, GEK\\
Writing—review \& editing: VK, JB, CNG, VO, GEK


\bibliographystyle{elsarticle-num} 
\begin{footnotesize}
\bibliography{references}

\begin{thebibliography}{10}
\expandafter\ifx\csname url\endcsname\relax
  \def\url#1{\texttt{#1}}\fi
\expandafter\ifx\csname urlprefix\endcsname\relax\def\urlprefix{URL }\fi
\expandafter\ifx\csname href\endcsname\relax
  \def\href#1#2{#2} \def\path#1{#1}\fi

\bibitem{Geometric_DL_Bronstein}
M.~M. Bronstein, J.~Bruna, Y.~LeCun, A.~Szlam, P.~Vandergheynst, {Geometric Deep Learning: Going beyond Euclidean data}, IEEE Signal Processing Magazine 34~(4) (2017) 18--42.
\newblock \href {https://doi.org/10.1109/MSP.2017.2693418} {\path{doi:10.1109/MSP.2017.2693418}}.

\bibitem{Deep_generative_learning_Faez}
L.~Regenwetter, A.~H. Nobari, F.~Ahmed, \href{https://doi.org/10.1115/1.4053859}{{Deep Generative Models in Engineering Design: A Review}}, Journal of Mechanical Design 144~(7) (2022) 071704.
\newblock \href {http://arxiv.org/abs/https://asmedigitalcollection.asme.org/mechanicaldesign/article-pdf/144/7/071704/6866682/md\_144\_7\_071704.pdf} {\path{arXiv:https://asmedigitalcollection.asme.org/mechanicaldesign/article-pdf/144/7/071704/6866682/md\_144\_7\_071704.pdf}}, \href {https://doi.org/10.1115/1.4053859} {\path{doi:10.1115/1.4053859}}.
\newline\urlprefix\url{https://doi.org/10.1115/1.4053859}

\bibitem{deeponet_lulu_2021}
L.~Lu, P.~Jin, G.~Pang, Z.~Zhang, G.~E. Karniadakis, {Learning nonlinear operators via DeepONet based on the universal approximation theorem of operators}, Nature Machine Intelligence 3~(3) (2021) 218--229.

\bibitem{FNO_2020}
Z.~Li, N.~Kovachki, K.~Azizzadenesheli, B.~Liu, K.~Bhattacharya, A.~Stuart, A.~Anandkumar, Fourier neural operator for parametric partial differential equations, arXiv preprint arXiv:2010.08895 (2020).

\bibitem{WNO}
T.~Tripura, S.~Chakraborty, Wavelet neural operator for solving parametric partial differential equations in computational mechanics problems, Computer Methods in Applied Mechanics and Engineering 404 (2023) 115783.

\bibitem{transolver}
H.~Wu, H.~Luo, H.~Wang, J.~Wang, M.~Long, {Transolver: a fast transformer solver for PDEs on general geometries}, in: Proceedings of the 41st International Conference on Machine Learning, 2024, pp. 53681--53705.

\bibitem{Geo-FNO}
Z.~Li, D.~Z. Huang, B.~Liu, A.~Anandkumar, Fourier neural operator with learned deformations for pdes on general geometries, Journal of Machine Learning Research 24~(388) (2023) 1--26.

\bibitem{meshgraphnet}
T.~Pfaff, M.~Fortunato, A.~Sanchez-Gonzalez, P.~Battaglia, Learning mesh-based simulation with graph networks, in: International Conference on Learning Representations, 2021.

\bibitem{Xmeshgraphnet_NVIDIA}
M.~A. Nabian, C.~Liu, R.~Ranade, S.~Choudhry, {X-meshgraphnet: Scalable multi-scale graph neural networks for physics simulation}, arXiv preprint arXiv:2411.17164 (2024).

\bibitem{drivaerml}
N.~Ashton, C.~Mockett, M.~Fuchs, L.~Fliessbach, H.~Hetmann, T.~Knacke, N.~Schonwald, V.~Skaperdas, G.~Fotiadis, A.~Walle, et~al., {DrivAerML: High-fidelity computational fluid dynamics dataset for road-car external aerodynamics}, arXiv preprint arXiv:2408.11969 (2024).

\bibitem{airfoil_shukla}
K.~Shukla, V.~Oommen, A.~Peyvan, M.~Penwarden, N.~Plewacki, L.~Bravo, A.~Ghoshal, R.~M. Kirby, G.~E. Karniadakis, \href{https://www.sciencedirect.com/science/article/pii/S0952197623017992}{Deep neural operators as accurate surrogates for shape optimization}, Engineering Applications of Artificial Intelligence 129 (2024) 107615.
\newblock \href {https://doi.org/https://doi.org/10.1016/j.engappai.2023.107615} {\path{doi:https://doi.org/10.1016/j.engappai.2023.107615}}.
\newline\urlprefix\url{https://www.sciencedirect.com/science/article/pii/S0952197623017992}

\bibitem{fusiondeeponet}
A.~Peyvan, V.~Kumar, G.~E. Karniadakis, {Fusion-DeepONet: A Data-Efficient Neural Operator for Geometry-Dependent Hypersonic and Supersonic Flows}, arXiv preprint arXiv:2501.01934 (2025).

\bibitem{synergistic_learning}
V.~Kumar, S.~Goswami, K.~Kontolati, M.~D. Shields, G.~E. Karniadakis, \href{https://www.sciencedirect.com/science/article/pii/S0893608024010426}{{Synergistic learning with multi-task DeepONet for efficient PDE problem solving}}, Neural Networks 184 (2025) 107113.
\newblock \href {https://doi.org/https://doi.org/10.1016/j.neunet.2024.107113} {\path{doi:https://doi.org/10.1016/j.neunet.2024.107113}}.
\newline\urlprefix\url{https://www.sciencedirect.com/science/article/pii/S0893608024010426}

\bibitem{geom_deeponet}
J.~He, S.~Koric, D.~Abueidda, A.~Najafi, I.~Jasiuk, {Geom-DeepONet: A point-cloud-based deep operator network for field predictions on 3D parameterized geometries}, Computer Methods in Applied Mechanics and Engineering 429 (2024) 117130.

\bibitem{pointnet}
C.~R. Qi, H.~Su, K.~Mo, L.~J. Guibas, {Pointnet: Deep learning on point sets for 3D classification and segmentation}, in: Proceedings of the IEEE conference on computer vision and pattern recognition, 2017, pp. 652--660.

\bibitem{pointnet++}
C.~R. Qi, L.~Yi, H.~Su, L.~J. Guibas, {Pointnet++: Deep hierarchical feature learning on point sets in a metric space}, Advances in neural information processing systems 30 (2017).

\bibitem{pointcnn}
Y.~Li, R.~Bu, M.~Sun, W.~Wu, X.~Di, B.~Chen, {Pointcnn: Convolution on x-transformed points}, Advances in neural information processing systems 31 (2018).

\bibitem{pointmlp}
X.~Ma, C.~Qin, H.~You, H.~Ran, Y.~Fu, Rethinking network design and local geometry in point cloud: A simple residual mlp framework, arXiv preprint arXiv:2202.07123 (2022).

\bibitem{mionet}
S.~Xiao, P.~Jin, Y.~Tang, {Learning solution operators of PDEs defined on varying domains via MIONet}, arXiv preprint arXiv:2402.15097 (2024).

\bibitem{point_deeponet}
J.~Park, N.~Kang, {Point-DeepONet: A Deep Operator Network Integrating PointNet for Nonlinear Analysis of Non-Parametric 3D Geometries and Load Conditions}, arXiv preprint arXiv:2412.18362 (2024).

\bibitem{GINO}
Z.~Li, N.~Kovachki, C.~Choy, B.~Li, J.~Kossaifi, S.~Otta, M.~A. Nabian, M.~Stadler, C.~Hundt, K.~Azizzadenesheli, et~al., {Geometry-informed neural operator for large-scale 3D PDEs}, Advances in Neural Information Processing Systems 36 (2023) 35836--35854.

\bibitem{CORAL}
L.~Serrano, L.~Le~Boudec, A.~Kassa{\"\i}~Koupa{\"\i}, T.~X. Wang, Y.~Yin, J.-N. Vittaut, P.~Gallinari, {Operator learning with neural fields: Tackling pdes on general geometries}, Advances in Neural Information Processing Systems 36 (2023) 70581--70611.

\bibitem{domino}
R.~Ranade, M.~A. Nabian, K.~Tangsali, A.~Kamenev, O.~Hennigh, R.~Cherukuri, S.~Choudhry, Domino: A decomposable multi-scale iterative neural operator for modeling large scale engineering simulations, arXiv preprint arXiv:2501.13350 (2025).

\bibitem{physics_geom_aware_Meidi}
W.~Zhong, H.~Meidani, Physics-informed geometry-aware neural operator, Computer Methods in Applied Mechanics and Engineering 434 (2025) 117540.

\bibitem{attention_transformers}
A.~Vaswani, N.~Shazeer, N.~Parmar, J.~Uszkoreit, L.~Jones, A.~N. Gomez, {\L}.~Kaiser, I.~Polosukhin, Attention is all you need, Advances in neural information processing systems 30 (2017).

\bibitem{GNOT}
Z.~Hao, Z.~Wang, H.~Su, C.~Ying, Y.~Dong, S.~Liu, Z.~Cheng, J.~Song, J.~Zhu, Gnot: A general neural operator transformer for operator learning, in: International Conference on Machine Learning, PMLR, 2023, pp. 12556--12569.

\bibitem{transolver++}
H.~Luo, H.~Wu, H.~Zhou, L.~Xing, Y.~Di, J.~Wang, M.~Long, {Transolver++: An Accurate Neural Solver for PDEs on Million-Scale Geometries}, arXiv preprint arXiv:2502.02414 (2025).

\bibitem{drivaernet++}
M.~Elrefaie, F.~Morar, A.~Dai, F.~Ahmed, Drivaernet++: A large-scale multimodal car dataset with computational fluid dynamics simulations and deep learning benchmarks, Advances in Neural Information Processing Systems 37 (2024) 499--536.

\bibitem{latentneuraloperator}
T.~Wang, C.~Wang, Latent neural operator for solving forward and inverse pde problems, Advances in Neural Information Processing Systems 37 (2024) 33085--33107.

\bibitem{adaptiveDOE_Dassault}
{Dassault Syst\`emes Corp}, \href{https://www.3ds.com/support/documentation/user-guides}{{SIMULIA User Assistance 2025, Adaptive DOE Technique}} (2025).
\newline\urlprefix\url{https://www.3ds.com/support/documentation/user-guides}

\bibitem{Abaqus_implicit}
{Dassault Syst\`emes Corp}, \href{https://www.3ds.com/support/documentation/user-guides}{{SIMULIA User Assistance 2025, Abaqus Analysis Guide}} (2025).
\newline\urlprefix\url{https://www.3ds.com/support/documentation/user-guides}

\bibitem{universal_approx}
T.~Chen, H.~Chen, Universal approximation to nonlinear operators by neural networks with arbitrary activation functions and its application to dynamical systems, IEEE transactions on neural networks 6~(4) (1995) 911--917.

\bibitem{pyvista}
B.~Sullivan, A.~Kaszynski, \href{https://doi.org/10.21105/joss.01450}{{PyVista}: {3D} plotting and mesh analysis through a streamlined interface for the {Visualization Toolkit} ({VTK})}, Journal of Open Source Software 4~(37) (2019) 1450.
\newblock \href {https://doi.org/10.21105/joss.01450} {\path{doi:10.21105/joss.01450}}.
\newline\urlprefix\url{https://doi.org/10.21105/joss.01450}

\bibitem{gelu}
D.~Hendrycks, K.~Gimpel, {Gaussian error linear units (GELUs)}, arXiv preprint arXiv:1606.08415 (2016).

\bibitem{onecycleLR}
L.~N. Smith, N.~Topin, Super-convergence: Very fast training of neural networks using large learning rates, in: Artificial intelligence and machine learning for multi-domain operations applications, Vol. 11006, SPIE, 2019, pp. 369--386.

\end{thebibliography}
\end{footnotesize}

\clearpage
\newpage
\makeatletter
\renewcommand \thesection{S\@arabic\c@section}
\renewcommand\thetable{S\@arabic\c@table}
\renewcommand \thefigure{S\@arabic\c@figure}
\makeatother

\end{document}